\documentclass[3p,times,procedia]{elsarticle}
\usepackage{ecrc}
\usepackage[bookmarks=false]{hyperref}
    \hypersetup{colorlinks,
      linkcolor=blue,
      citecolor=blue,
      urlcolor=blue}
\usepackage{amsmath}
\usepackage{amssymb}
\usepackage{graphicx}
\usepackage[figuresright]{rotating}

\graphicspath{{figures/}}
\biboptions{numbers}

\volume{00}

\firstpage{1}

\journalname{Procedia Computer Science}

\runauth{G.F. Preziosa et al.}

\jid{procs}

\begin{document}
\begin{frontmatter}

\dochead{8th International Conference on Industry of the Future and Smart Manufacturing}

\title{Anomaly Detection on Small Industrial Components via Vision-Based Tactile Sensing}

\author[First]{Giuseppe Fabio Preziosa}
\author[First]{Matilde Casiglia}
\author[First]{Marco Faroni}
\author[First]{Andrea Maria Zanchettin}
\author[First]{Paolo Rocco}

\address[First]{The authors are with Politecnico di Milano, Piazza Leonardo da Vinci, 32, Milano (Italy). e-mail: (giuseppefabio.preziosa, matilde.casiglia, marco.faroni, andreamaria.zanchettin, paolo.rocco)@polimi.it}

\vspace*{-3pt}
\begin{center}
{\footnotesize\itshape Accepted at ISM 2026, Padua \& Venice, Italy, November 10--14, 2026.}
\end{center}
\vspace*{-9pt}

\begin{abstract}
Automated inspection of small industrial components --- including sub-centimetre-scale parts where defects are geometry-driven and poorly resolved by standard optical cameras --- calls for sensing modalities that can directly capture fine surface geometry. Vision-based tactile sensors address this need by converting contact imprints into high-resolution image-like data compatible with existing deep-learning pipelines, yet their effective use for industrial anomaly detection (AD) remains largely unexplored. This work systematically evaluates unsupervised AD methods on a real tactile dataset covering five genuine industrial components acquired with a GelSight Mini sensor mounted on a collaborative robot. Four feature-embedding methods --- SPADE, PaDiM, FAPM, and InReaCh --- are compared under three validations explicitly motivated by the deployment constraints of contact-based sensing: a Good Fraction analysis establishing the minimum number of nominal contacts for stable performance, directly bounded by gel wear since every acquisition degrades the soft interface; a cross-position evaluation assessing generalization across different contact locations observing the same recurring surface pattern; and a low- versus high-resolution comparison evaluating the cost--benefit of higher-resolution tactile acquisition. Overall, this systematic benchmarking study provides practical guidance for researchers and practitioners adopting vision-based tactile sensing for industrial AD and shows how this modality can serve as a viable alternative for industrial quality-control tasks.
\end{abstract}

\end{frontmatter}

\vspace*{-6pt}

\section{Introduction}
\label{sec:introduction}

Industrial inspection is increasingly expected to support flexible production systems, where product quality must be monitored with limited downtime, limited manual intervention, and reliable decisions under changing operating conditions. In this context, automated quality control must handle subtle defects, changing batches, and strict false-alarm requirements, because false positives interrupt production while missed defects may propagate to assembly, warranty, or safety-critical stages \cite{cui2023_survey_unsupervised_industrial_images}.

Although a fully supervised inspection framework could be considered, collecting defective parts, assigning them to fault categories, and building a representative training set is often impractical in industrial production. Defects are infrequent, can appear with heterogeneous morphologies, and are seldom available in sufficient quantity, while nominal production data are comparatively easier to acquire. This is the setting addressed by industrial anomaly detection (AD), which refers to the identification of deviations from a nominal condition in data acquired from products, processes, or machines. Rather than assuming fixed defect classes, AD has therefore developed largely around unsupervised or nominal-only formulations: many pipelines model normality from non-defective observations and, at test time, assess how much a sample departs from that reference model. Depending on the inspection requirement, the output may be an item-level decision or a spatially resolved anomaly-score map used for defect localization \cite{tao2022_unsup_anom_localization_survey}.

Most industrial anomaly-detection methods are developed on images, and in particular on RGB images. This is reflected by the most common benchmarks in the field, such as MVTec AD \cite{bergmann2021mvtec}, and by survey evidence showing that deep learning has enabled a broad family of algorithms for both image-level detection and pixel-level localization \cite{lin2024_survey_rgb_3d_multimodal_uiad}. Building on this progress, modern methods can often combine nominal-only training with pretrained representations, making them attractive when complete defect taxonomies are unavailable.

However, when parts are very small and the relevant defects are mainly geometric, the discriminative signal may be weak in standard RGB images. Small burrs, dents, local flattening, thread irregularities, and edge damage can be hard to observe reliably without specialized optics, controlled illumination, or dedicated metrology equipment. Manual inspection remains common in these settings, but it is time-consuming and affected by attention, fatigue, and subjectivity. Dedicated inspection machines can provide accurate measurements, but they are often expensive and less flexible when part geometries or production needs change.
Among the possible alternatives, vision-based tactile sensors are attractive because they preserve the image-like nature of the data while changing what the image represents: instead of recording optical appearance, they capture a contact imprint related to local geometry and surface relief. GelSight-type sensors can resolve details at the $\mu$m scale by imaging the deformation of a soft elastomer under controlled illumination, and compact devices such as the GelSight Mini make this modality accessible for robotic acquisition \cite{yuan2017gelsight}. In scenarios where inspection can be focused on a specific region of interest, vision-based tactile sensors offer a localized and complementary alternative: they can provide a relatively cost-effective inspection tool compared to dedicated metrology machines, while also aiming to reduce the operator-dependent subjectivity that affects purely manual visual inspections. Building on this motivation, this paper presents the following contributions:
\begin{itemize}
\item This work provides the first systematic demonstration and evaluation of vision-based tactile sensing for industrial anomaly detection on small real-world industrial components, offering practical guidance on how these sensors can be used for anomaly detection and quality-control tasks.
\item We compare unsupervised AD methods under tactile-specific constraints, focusing on the number of nominal images to acquire, generalization across acquisition positions that observe the same recurrent pattern, and the resolution to adopt for low- versus high-resolution tactile inspection.
\end{itemize}

\section{Related Work}
\label{sec:related_work}

Industrial AD provides the methodological setting for detecting departures from nominal production data, while the present study focuses on how this setting can be applied when data are acquired through vision-based tactile sensing. Three elements are therefore useful to position the work: the sensor (Section~\ref{subsec:related_tactile_sensors}), the available benchmarks (Section~\ref{subsec:related_datasets}), and the detection methods (Section~\ref{subsec:related_methods}).

\subsection{Vision-Based Tactile Sensors}
\label{subsec:related_tactile_sensors}
Vision-based tactile sensors acquire information through physical contact, producing RGB images that describe local surface details in the contact area. GelSight pioneered this approach by showing how a camera observing a deformable elastomer from behind can yield high-resolution tactile images \cite{yuan2017gelsight}; DIGIT later made the same principle accessible through a compact, low-cost design \cite{lambeta2020digit}. These sensors have been used mainly for robotic manipulation tasks, including grasp-outcome prediction \cite{calandra2017feeling}, gentle grasping of deformable objects \cite{han2025_vision_tactile_grasping}, and slip detection \cite{li2018_slip_detection}. However, their industrial adoption remains less established because contact-based sensing introduces specific limitations: soft gel interfaces wear with repeated contacts, acquisition throughput depends on the contact cycle, and small sensing areas require carefully selected regions of interest. Recent reviews and sensor developments therefore identify durability, sensing-area scalability, and deployment robustness as key issues for real-world use \cite{li2025vbtsreview,mirzaee2025gelbelt}. In this work, these physical characteristics motivate an ad hoc evaluation of tactile sensing, including the number of images, and therefore contacts, required to model nominality and the image resolution to use for defect localization.

\subsection{Datasets for Unsupervised Industrial Anomaly Detection}
\label{subsec:related_datasets}
Benchmarking in unsupervised industrial AD relies on RGB image datasets with nominal-only training and defective test samples. MVTec AD \cite{bergmann2021mvtec} and VisA \cite{zou2022spotdiff} are the primary references, while KolektorSDD \cite{tabernik2019jim} addresses a real surface-defect production case. All share the same dominant modality---optical RGB imaging---and evaluate appearance or texture deviations rather than contact-derived geometric signatures. No established tactile dataset exists for unsupervised industrial AD on small real-world components, leaving modern localization methods untested under tactile acquisition constraints.

\subsection{Anomaly Detection Methods}
\label{subsec:related_methods}
Unsupervised industrial AD methods can be grouped according to how they represent nominality and how anomaly evidence is extracted at test time. 

A first family is based on feature embeddings: a pretrained network maps nominal images, or local patches, into a latent space where normality can be represented through boundaries, distributions, or stored exemplars. This family includes one-class methods, Teacher--Student methods, distribution-modeling methods, and memory-bank methods. One-class methods learn compact latent descriptions of nominal samples and classify by distance from the learned boundary, with PatchSVDD \cite{yi2020_patch_svdd} extending this principle from image-level descriptors to patch-level localization. Teacher--Student methods use distillation instead: a student network is trained to reproduce the features of a fixed teacher on nominal data, and anomalous regions are detected where teacher and student responses diverge, as in STPM \cite{wang2021stfpm}. Distribution-modeling methods replace explicit boundaries with local statistical models. PaDiM \cite{defard2020padim} fits, at each spatial location of a pretrained feature grid, a multivariate Gaussian estimated from nominal patches and scores anomalies through the Mahalanobis distance. Memory-bank methods represent normality more explicitly by storing nominal embeddings and comparing test descriptors by nearest-neighbor search. SPADE \cite{cohen2020spade} builds a memory of normal features at multiple pyramid scales and localizes anomalies through deep correspondences between a test image and the nearest nominal exemplars. FAPM \cite{xu2022fapm} further improves efficiency through adaptive patch selection and co-located prototype banks, whereas InReaCh \cite{mcintosh2023inreach} exploits inter-realization channel consistency to retain repeatable nominal correspondences and filter rare patterns before scoring. These feature-embedding approaches are attractive for industrial deployment because they require little or no training beyond feature extraction, but their performance depends on how well the nominal representation covers acquisition variability.

A second family comprises reconstruction-based methods: a model trained on nominal samples is applied at test time and anomalies are detected as residuals between input and reconstruction, exploiting the assumption that anomalous patterns---absent during training---will be reproduced less faithfully. Implementations range from autoencoder-based inpainting, as in RIAD \cite{zavrtanik2021riad}, to adversarial generative models, as in DC-AE \cite{zhang2024dcae}. These methods can be powerful but typically require more training effort and may partially reconstruct anomalous content when the model generalizes too broadly.

A third family is represented by synthesized-anomaly methods, which address the lack of real defects by generating artificial anomalies during training. CutPaste \cite{li2021cutpaste} creates local cut-and-paste irregularities and learns defect-sensitive representations from the resulting pretext task, whereas DRAEM \cite{zavrtanik2021draem} synthesizes anomaly masks and trains a reconstruction-discrimination pipeline for dense localization. Such methods introduce useful supervision without real defective samples. 

In light of this overview, this work focuses on feature-embedding methods because they suit the practical constraints of tactile inspection. Compared with methods that must learn reconstructions or defect-sensitive transformations from scratch, they typically require fewer nominal training images, reducing the number of contacts and the associated gel wear. This is also important for rapid adaptation to new components, a key requirement in modern industrial production, where collecting extensive tactile data for each new geometry would limit deployment flexibility.

\section{Tactile Inspection Dataset}
\label{sec:dataset}

This section describes the tactile inspection dataset used in the study. Section~\ref{subsec:dataset_context} introduces the industrial components and inspection targets, Section~\ref{subsec:defect_generation} describes how defective samples are generated, Section~\ref{subsec:acquisition} details the acquisition hardware and protocol, and Section~\ref{subsec:composition} summarizes the resulting dataset composition and annotations.

\subsection{Industrial Context and Component Overview}
\label{subsec:dataset_context}

\begin{figure}[t]\vspace*{4pt}
\centering
\includegraphics[width=\linewidth]{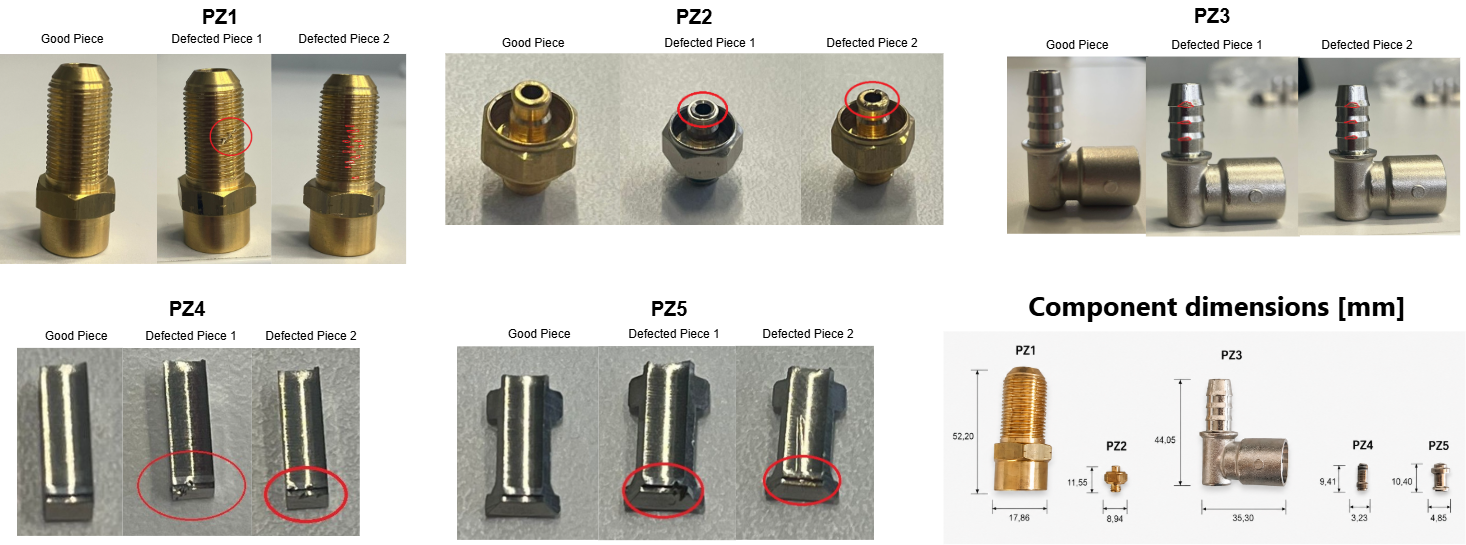}
\caption{Overview of the five inspected industrial components and their reference dimensions.}
\label{fig:components_overview}
\end{figure}

The dataset consists of five industrial components, referred to as PZ1--PZ5, selected according to two criteria: they are real components for which inspection is genuinely challenging in practice, typically requiring manual checks, dedicated tools, gauges, or ad hoc machines with limited flexibility; and they provide heterogeneity in both size and defect typology, enabling evaluation across geometries, scales, and defect types. As shown in Fig.~\ref{fig:components_overview}, the components range from threaded or grooved parts to sub-centimetre-scale components.
(i)~\textbf{PZ1} is a threaded component; inspection targets profile regularity, thread continuity, and edge sharpness, which may be degraded by flattening, local irregularities, and discontinuities.
(ii)~\textbf{PZ2} is a small connector-like component whose critical region is a high-tolerance fitting interface, where even small dents, marks, or edge irregularities may affect functional coupling.
(iii)~\textbf{PZ3} contains repetitive geometric features (grooves, slots, and a tooth-like feature verified with a dedicated gauge); inspection targets groove edge sharpness and border regularity.
(iv)~\textbf{PZ4} and \textbf{PZ5} are sub-centimetre-scale components whose reduced size makes inspection highly sensitive to small dents, scratches, burrs, and local material loss.

\subsection{Defect Generation}
\label{subsec:defect_generation}

Defects are generated artificially with the explicit aim of reproducing damage conditions similar to those encountered in real industrial processing and handling scenarios. The defect-generation process is supervised by a domain-expert engineer to ensure that the resulting damage remains representative of plausible industrial defects. Different tools, including pliers, hammers, and sandpaper, are used to induce heterogeneous damage mechanisms and tactile-visual signatures, covering plastic deformation, edge damage, and abrasion-like marks.
In addition, a subset of samples is subjected to controlled drop or impact events in order to emulate handling-related damage that may occur along production and logistics flows, for example when small parts are moved through hoppers, feeders, or gravity-based handling systems. Whenever applicable, defect severity is progressively increased up to functional failure. For PZ1, functional impairment is assessed through a mating test with a nut: the defect is considered true when the nut can no longer be screwed reliably. For PZ3--PZ5, even minimal defects are typically sufficient in practice to render the part unacceptable. PZ2 also includes genuine industrial defects that were not manually generated.

\subsection{Acquisition Hardware and Protocol}
\label{subsec:acquisition}

\begin{figure}[t]\vspace*{4pt}
\centering
\setlength{\tabcolsep}{1pt}
\renewcommand{\arraystretch}{1.0}
\begin{tabular}{@{}cccccc@{}}
\includegraphics[height=0.155\linewidth]{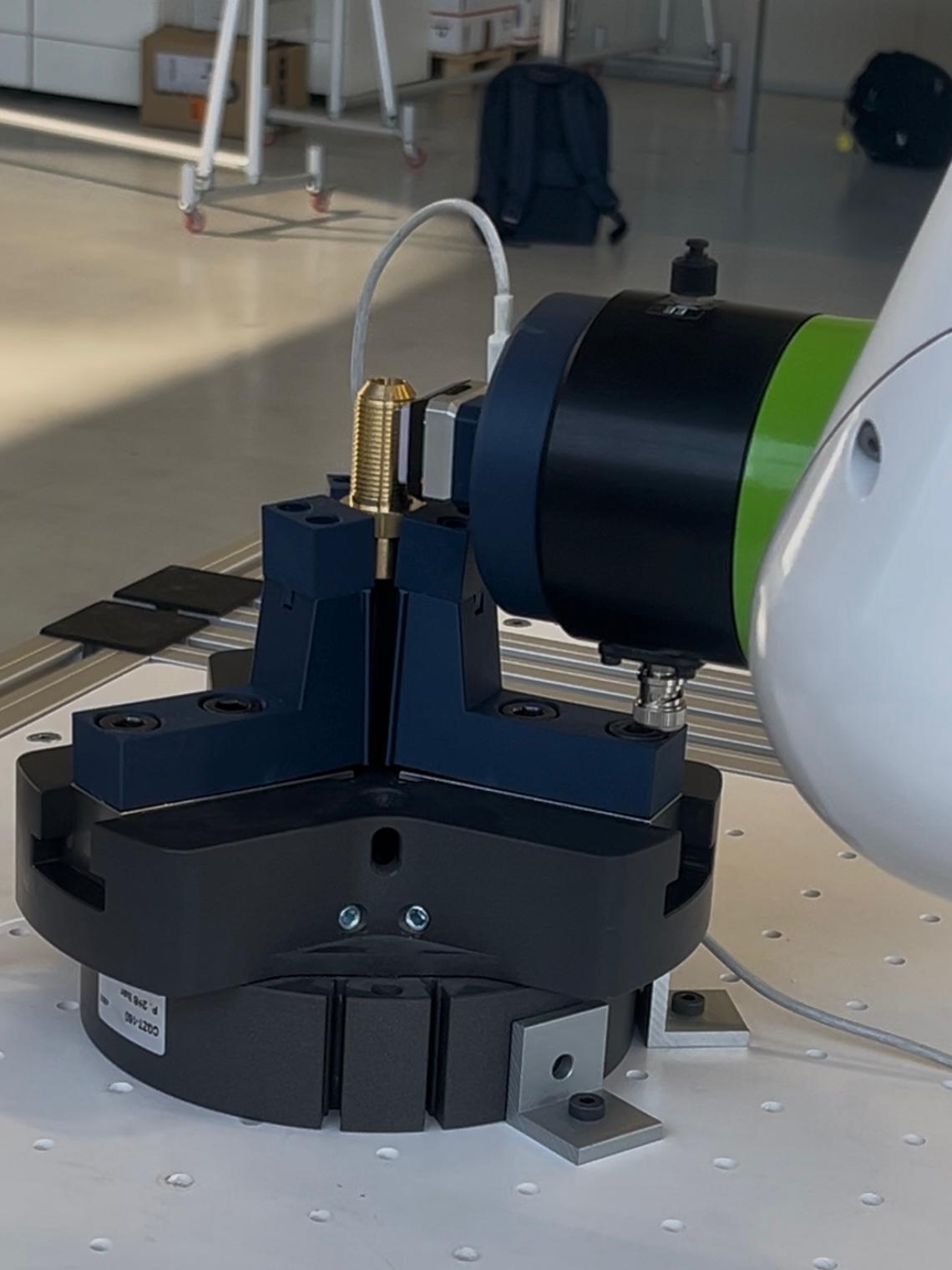} &
\includegraphics[height=0.155\linewidth]{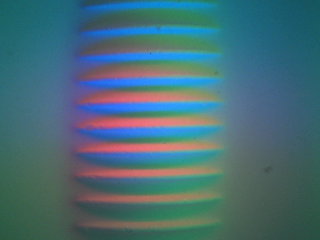} &
\includegraphics[height=0.155\linewidth]{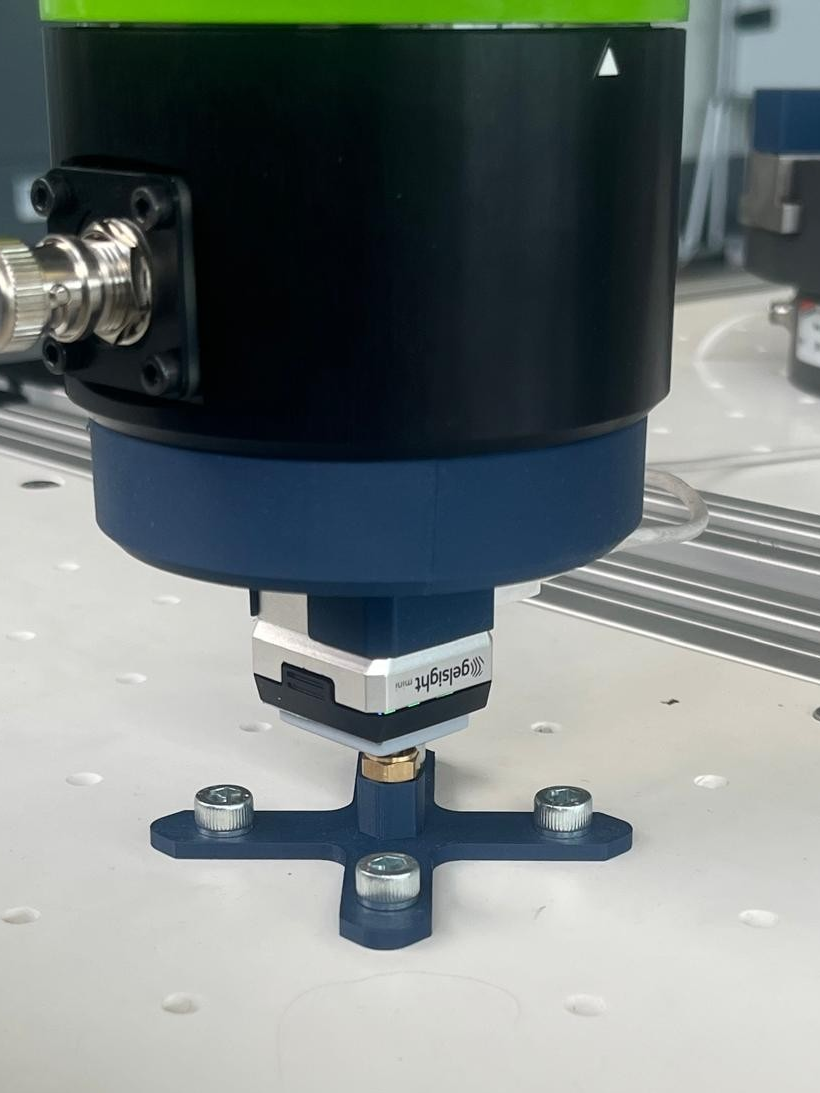} &
\includegraphics[height=0.155\linewidth]{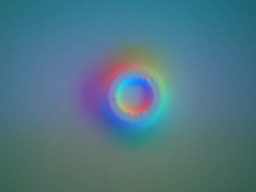} &
\includegraphics[height=0.155\linewidth]{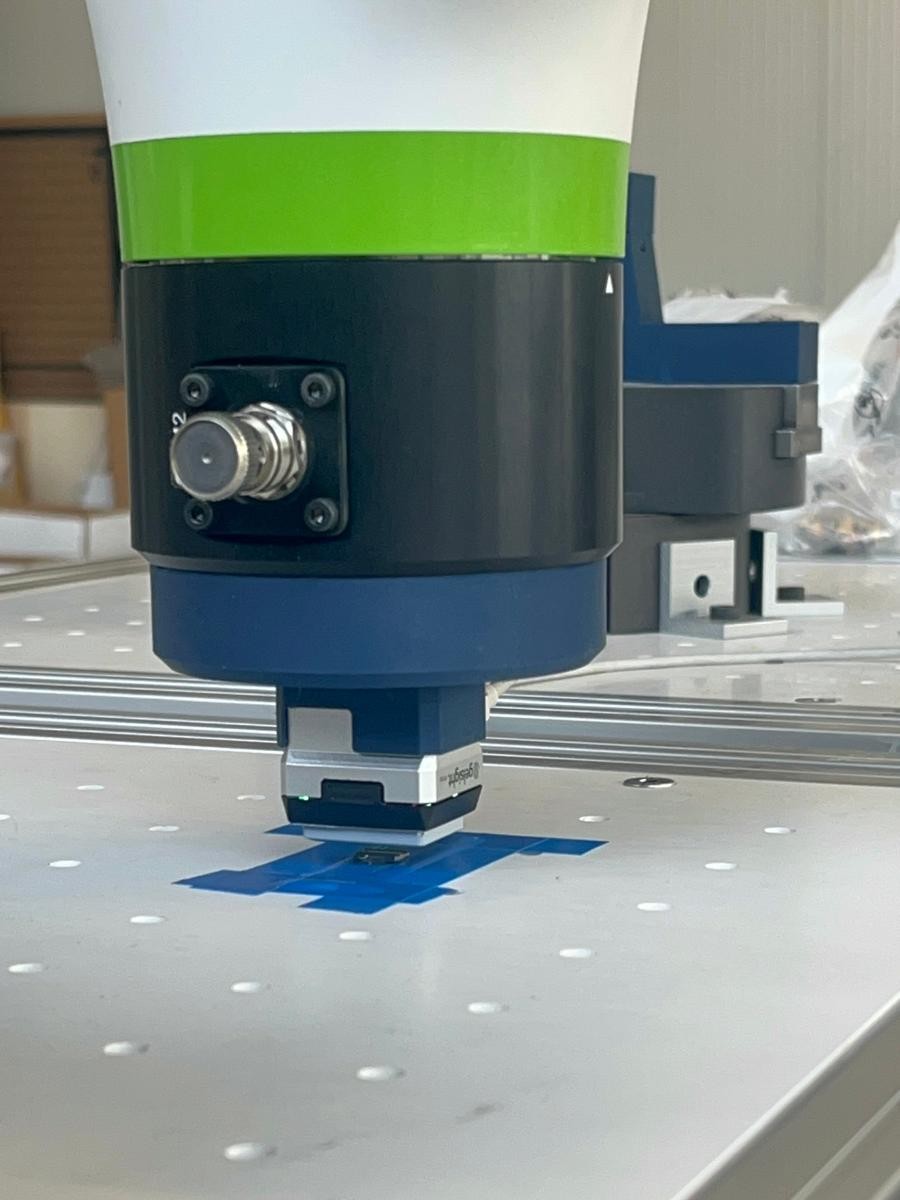} &
\includegraphics[height=0.155\linewidth]{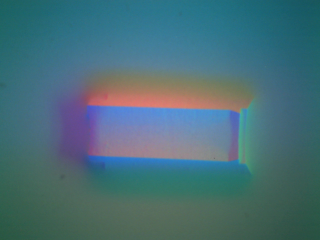} \\
\scriptsize Fixture PZ1 &
\scriptsize PZ1 frame &
\scriptsize Fixture PZ2 &
\scriptsize PZ2 frame &
\scriptsize Fixture PZ4--PZ5 &
\scriptsize PZ5 frame
\end{tabular}
\caption{Representative acquisition configurations and corresponding tactile frames.}
\label{fig:setup_vs_tactile_examples}
\end{figure}

Dataset acquisition is carried out using a GelSight Mini tactile sensor \cite{yuan2017gelsight}. The sensor is rigidly mounted on the wrist of a FANUC CRX-10iA collaborative robot through a dedicated flange, and robot motion is executed under position control. PZ1, PZ2, and PZ3 are mounted on 3D-printed fixtures to reduce placement variability, whereas PZ4 and PZ5 are placed manually, reflecting a more challenging and partly more realistic condition for sub-centimetre-scale components that are difficult to rigidly constrain. Over the complete acquisition campaign, two GelSight Mini gels were consumed, confirming that soft-interface wear is not only a theoretical deployment concern.
For each position, multiple acquisitions are performed by pressing, lifting, and re-pressing the sensor at the same nominal location. This protocol is particularly relevant in tactile inspection because each contact introduces small variations in positioning, alignment, and contact conditions, especially when simple 3D-printed fixtures are used. Representative acquisition configurations and corresponding tactile frames are shown in Fig.~\ref{fig:setup_vs_tactile_examples}.

\subsection{Dataset Composition}
\label{subsec:composition}

The dataset is composed of tactile images acquired from the five components, with two main acquisition variables: image resolution and number of acquisition positions per component. All components are acquired at $320\times240$, while PZ3, PZ4, and PZ5 are additionally acquired at a higher resolution of $1280\times720$.
Regarding acquisition positions, PZ1 uses four poses to sample the thread around the component, PZ3 uses two poses to cover the groove/tooth region, and PZ2, PZ4, and PZ5 use a single pose focused on a localized inspection area. As described in Section~\ref{subsec:acquisition}, each selected contact position is repeated ten times to capture contact variability under nominally identical robot commands. Table~\ref{tab:dataset_summary_paper} summarizes the dataset composition.
Defective regions are annotated using CVAT (Computer Vision Annotation Tool), producing binary ground-truth masks \cite{cvat_zenodo}. Annotations are performed by the same operator who generated the defects, ensuring consistency and direct knowledge of defect location and morphology. For PZ3--PZ5, ground-truth masks are first generated at native high resolution and then mapped to $320\times240$ through block-wise max pooling, preserving defect presence within each downsampling block.
The dataset is not released publicly due to industrial confidentiality constraints, but its structure, acquisition protocol, and evaluation splits are fully described.

\begin{table}[t]
\caption{Dataset composition for PZ1--PZ5. LR: $320\times240$; HR: $1280\times720$; Pos.: number of acquisition positions for the same component.}
\label{tab:dataset_summary_paper}
\vspace{3pt}
\centering
\scriptsize
\setlength{\tabcolsep}{2pt}
\renewcommand{\arraystretch}{0.95}
\resizebox{0.75\hsize}{!}{%
\begin{tabular}{lllllll}
\hline
Part & Description / ROI & Resolution & Fixture & Pos. & Normal & Defective \\
\hline
PZ1 & \begin{tabular}[c]{@{}l@{}}External thread; profile continuity\\and edge sharpness\end{tabular} & LR & Yes & 4 & 1600 & 75 \\
PZ2 & \begin{tabular}[c]{@{}l@{}}Upper fitting/adapter interface;\\high-tolerance fit region\end{tabular} & LR & Yes & 1 & 210 & 17 \\
PZ3 & \begin{tabular}[c]{@{}l@{}}Grooves/slots and tooth feature;\\edge regularity\end{tabular} & LR + HR & Yes & 2 & 260 & 24 \\
PZ4 & \begin{tabular}[c]{@{}l@{}}Sub-centimetre-scale component; edge sharpness\\and small defects\end{tabular} & LR + HR & No & 1 & 220 & 21 \\
PZ5 & \begin{tabular}[c]{@{}l@{}}Sub-centimetre-scale component; edge sharpness\\and small defects\end{tabular} & LR + HR & No & 1 & 230 & 21 \\
\hline
\end{tabular}%
}
\end{table}

\section{Implemented Feature-Embedding Methods}
\label{sec:implementation}

This work compares four methods from the feature-embedding family: SPADE \cite{cohen2020spade}, PaDiM \cite{defard2020padim}, FAPM \cite{xu2022fapm}, and InReaCh \cite{mcintosh2023inreach}. The considered approaches share a common principle: a pretrained backbone extracts deep features from nominal (``good'') samples, which are used to define a reference model of normality in feature space.

\begin{itemize}
\item \textbf{SPADE} stores nominal global descriptors and multi-scale local descriptors. At image level, each test sample is compared with the 5 nearest nominal images; localization is then obtained from the minimum patch-level distance to the nearest local nominal descriptor.
\item \textbf{PaDiM} models each spatial location with a multivariate Gaussian. Features are extracted from the last blocks of the first three backbone stages and concatenated into a 1792-dimensional descriptor; 550 channels are retained for scoring, and a regularized covariance estimate is used before computing Mahalanobis distances.
\item \textbf{FAPM} builds co-located prototype memories for each spatial location. Nominal descriptors are compressed into a near prototype bank retaining 10\% of the local descriptors, while more variable locations can use an expanded far bank with 2 times the prototype budget, corresponding to 20\% of the local descriptors.
\item \textbf{InReaCh} constructs inter-realization channels through mutual nearest-neighbor correspondences. Channels are built from 10 nominal realizations using a strict mutual nearest-neighbor criterion with Euclidean distance, then pruned by retaining only channels with at least 3 elements and internal dispersion below 5.0.
\end{itemize}

All methods use a Wide-ResNet-50-2 backbone pretrained on ImageNet and kept frozen. Low-resolution tactile images are processed as $224\times224$ central crops from the $320\times240$ acquisitions. For high-resolution PZ3--PZ5 images, a divide-and-conquer strategy \cite{rolih2024divideconquer} decomposes each $1280\times720$ image into overlapping $224\times224$ tiles with 56-pixel overlap and recomposes tile heatmaps through weighted blending.

\section{Experimental Comparisons}
\label{sec:experiments}

\subsection{Metrics}
\label{subsec:metrics}

Evaluation is performed at image and pixel level using threshold-free indicators, following industrial AD practice \cite{bergmann2019mvtec,bergmann2021mvtec}. AUROC is used at both levels: at image level each acquisition receives a scalar anomaly score, whereas at pixel level each dense anomaly map is compared against the binary ground-truth mask. Since defective pixels represent only a small fraction of the image, pixel-level evaluation also reports AUPRC, which is more informative under strong class imbalance. Finally, region-level localization is assessed with $\mathrm{AUC\text{-}PRO}@0.01$, which measures coverage of ground-truth defect regions while limiting background activation, integrated over $\mathrm{FPR}\leq0.01$ \cite{bergmann2019mvtec,bergmann2021mvtec}.

\subsection{Validation 1: Good Fraction Analysis}
\label{subsec:val1}

\begin{figure}[t]\vspace*{4pt}
\centering
\setlength{\tabcolsep}{1pt}
\renewcommand{\arraystretch}{1.0}
\begin{tabular}{@{}cccc@{}}
\includegraphics[width=0.248\linewidth]{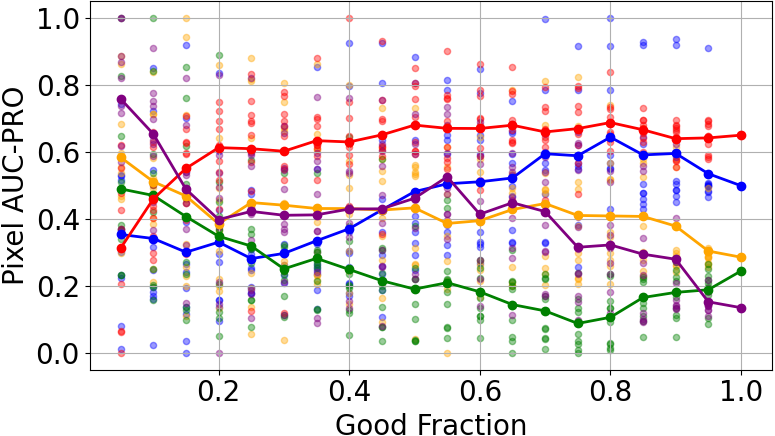} &
\includegraphics[width=0.248\linewidth]{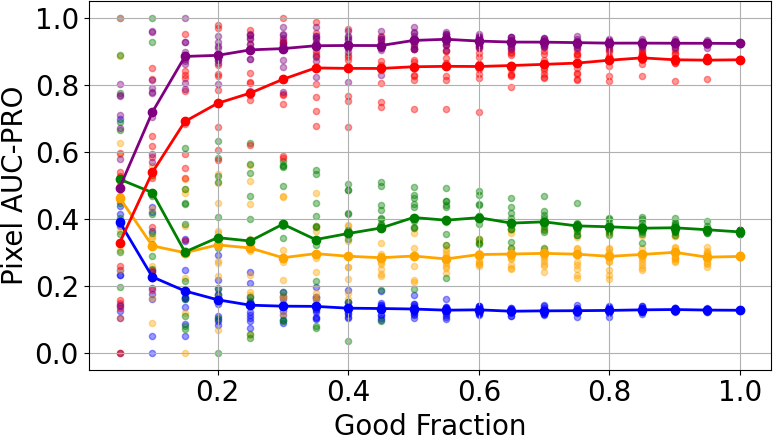} &
\includegraphics[width=0.248\linewidth]{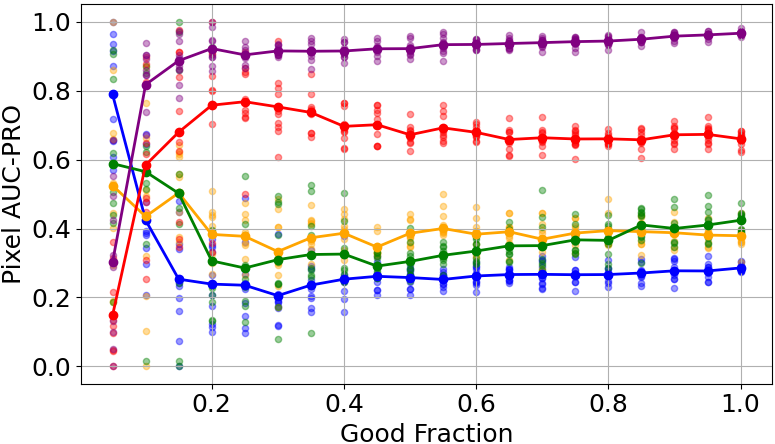} &
\includegraphics[width=0.248\linewidth]{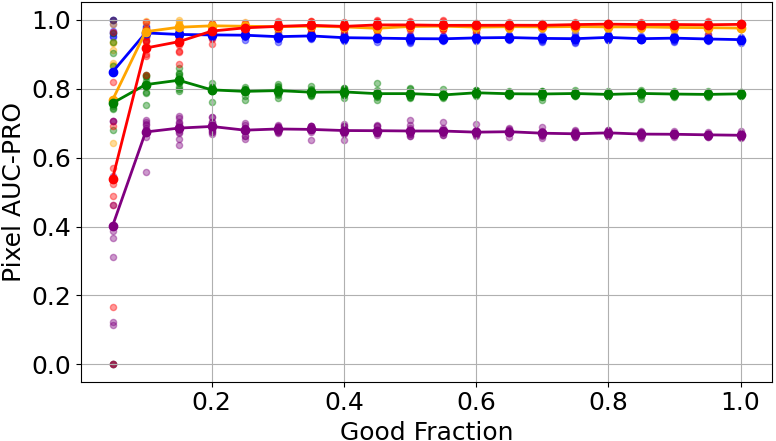} \\[1pt]
{\scriptsize (a) SPADE} &
{\scriptsize (b) PaDiM} &
{\scriptsize (c) FAPM} &
{\scriptsize (d) InReaCh} \\
\end{tabular}
\caption{Validation 1: normalized $\mathrm{AUC\text{-}PRO}@0.01$ curves vs.\ Good Fraction for each method. PZ1 (blue), PZ2 (yellow), PZ3 (green), PZ4 (red), PZ5 (purple). Dots are individual seeds; lines trace the per-component mean.}
\label{fig:val1_pro}
\end{figure}

\begin{figure*}[t]\vspace*{4pt}
\centering
\setlength{\tabcolsep}{2pt}
\renewcommand{\arraystretch}{1.0}
\begin{tabular}{@{}ccc@{}}
\includegraphics[width=0.325\textwidth]{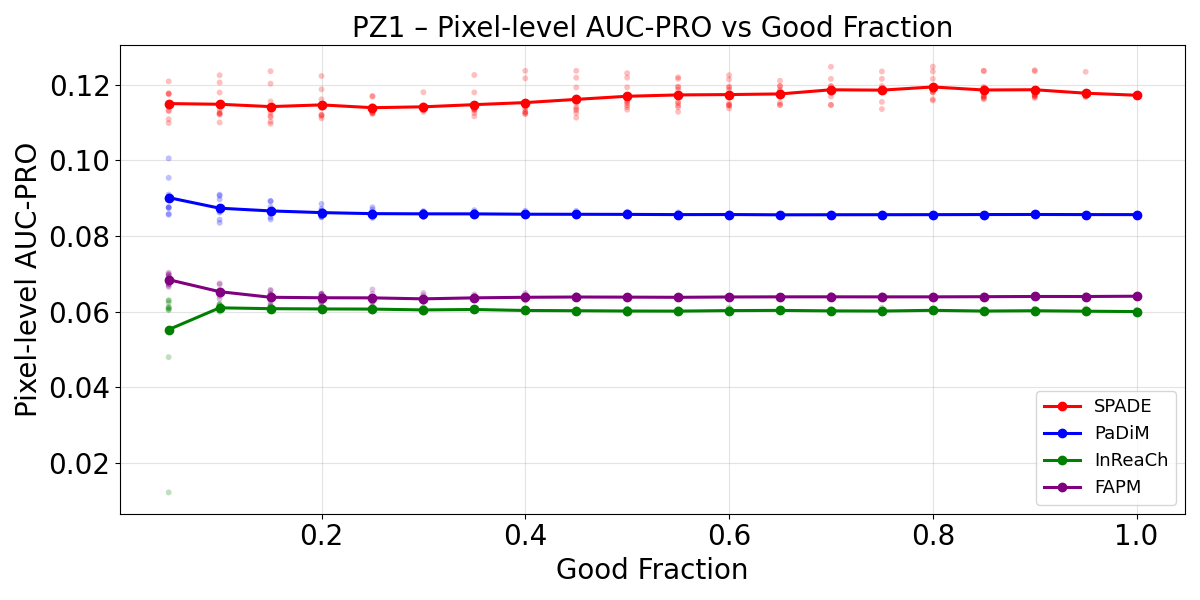} &
\includegraphics[width=0.325\textwidth]{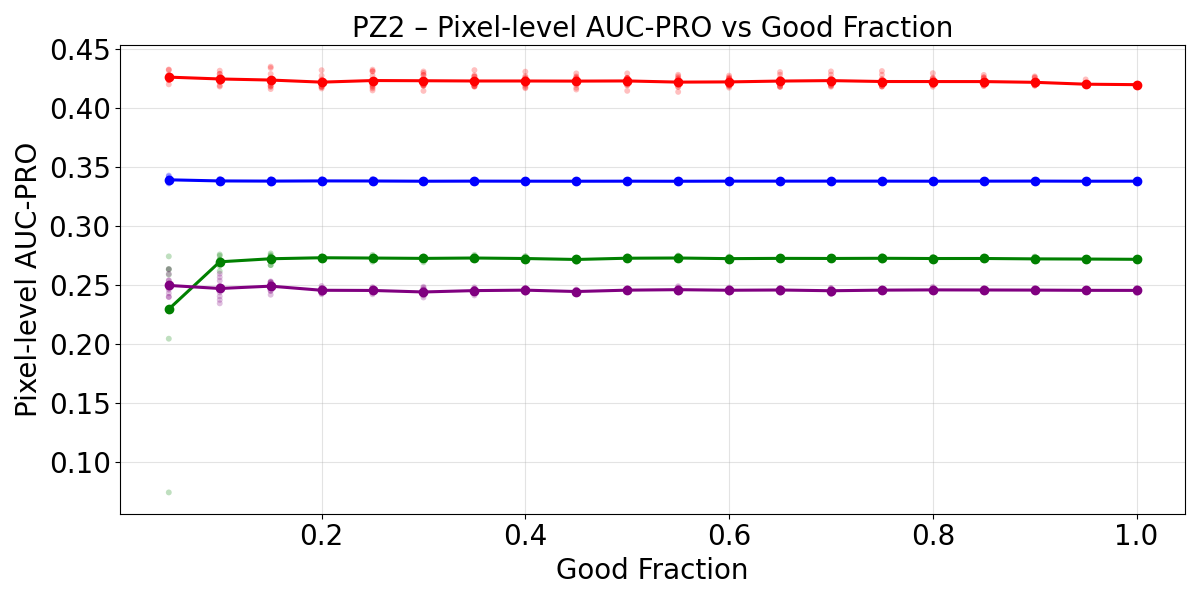} &
\includegraphics[width=0.325\textwidth]{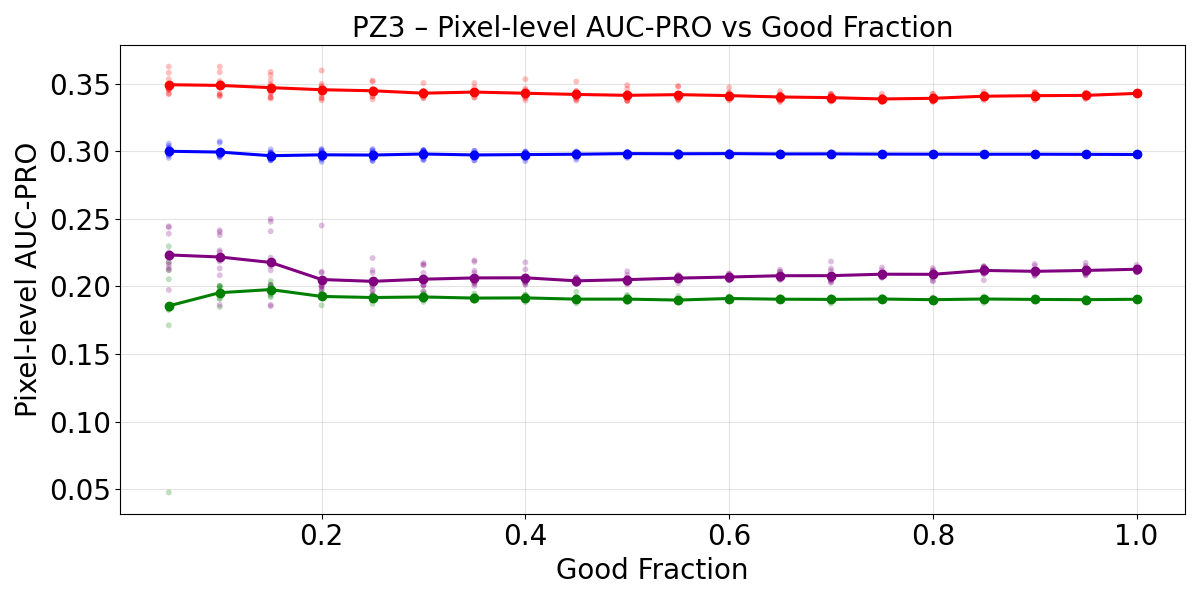} \\[-2pt]
\multicolumn{3}{c}{%
  \includegraphics[width=0.325\textwidth]{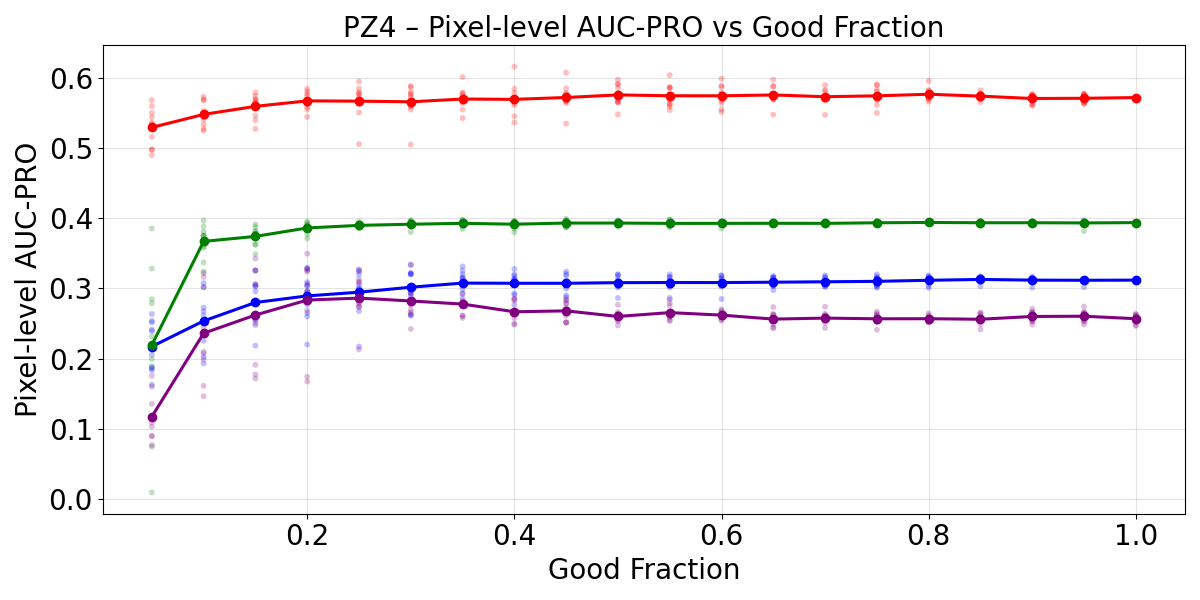}\hspace{4pt}%
  \includegraphics[width=0.325\textwidth]{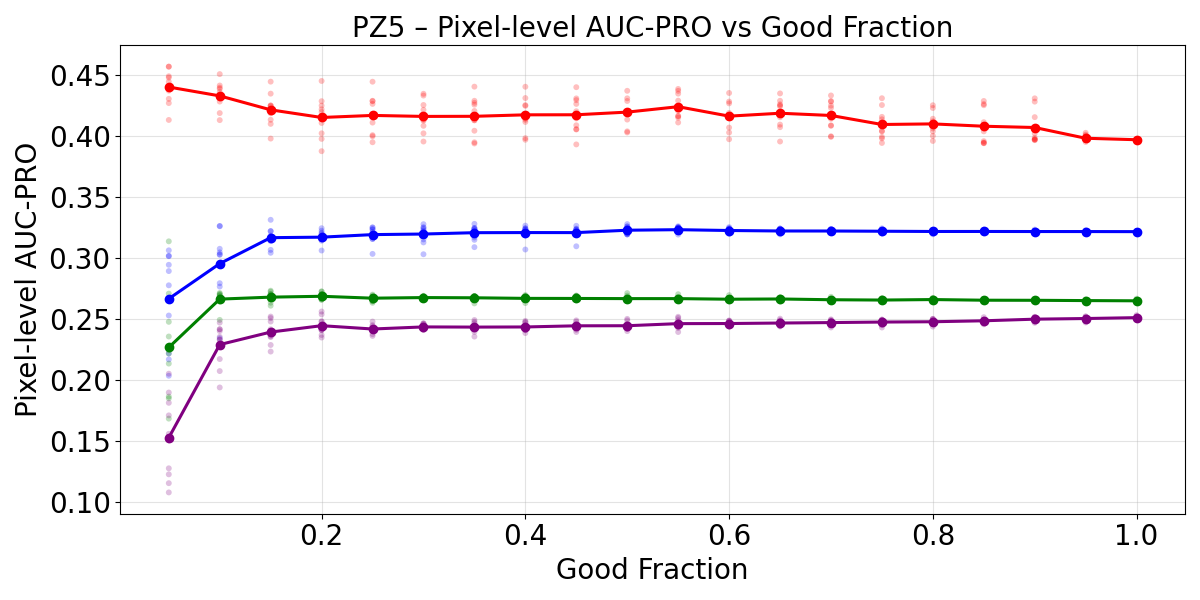}} \\
\end{tabular}
\caption{Validation 1: AUC-PRO@0.01 vs.\ Good Fraction for each component, all four methods plotted together (SPADE: red, PaDiM: blue, InReaCh: green, FAPM: purple). SPADE consistently attains the highest localization score across all five components. The gap is most pronounced on the sub-centimetre-scale components PZ4 and PZ5.}
\label{fig:val1_allpro}
\end{figure*}

Validation 1 studies how much nominal training data is needed before performance stabilizes. For each component, one acquisition position is selected as reference; 20 nominal images are held out for validation; the remaining nominal samples form the training pool; and defective validation samples are taken from the same position. The Good Fraction varies from 5\% to 100\% in 5\% steps, and each condition is repeated over 10 training seeds so that seed dispersion reflects sensitivity to the particular nominal subset. The stable Good Fraction is selected as a single method-level value, not separately for each component. For each method, we choose the smallest fraction after which the mean performance curves show a clear and consistent plateau across PZ1--PZ5 and across the metrics introduced in Section~\ref{subsec:metrics}. Fig.~\ref{fig:val1_pro} reports the normalized $\mathrm{AUC\text{-}PRO}@0.01$ curves; although only this indicator is plotted for compactness, it is representative of the trends observed for the other metrics. SPADE does not exhibit a single robust plateau, consistently with its matching-based formulation: the method relies on nearest-neighbor correspondences to nominal exemplars, so the effective coverage of the nominal set may keep changing as more samples are added, without necessarily producing a clean plateau in the present tactile setting. Therefore, the full nominal pool is retained for SPADE (100\%, 380/190/110/200/210 images for PZ1--PZ5, respectively). The other methods instead show a clearer threshold beyond which adding nominal samples does not provide systematic gains: PaDiM stabilizes around 40\% (152/76/44/80/84 images), FAPM around 60\% (228/114/66/120/126 images), and InReaCh around 20\% (76/38/22/40/42 images), in the same PZ1--PZ5 order. Interestingly, in some cases performance decreases when more nominal samples are added, and this non-monotone behavior is also observed in the other metrics, not only in $\mathrm{AUC\text{-}PRO}@0.01$. For prototype-memory methods such as FAPM, a plausible explanation is that the patch-wise coreset memory covers the nominal feature space and scores test features through Euclidean distances to the nearest nominal prototypes, but the resulting heatmap is not intrinsically calibrated for patch-specific distance scales. Patch-dependent normalization may therefore be needed to account for local feature variability.
Beyond the convergence behavior, Validation 1 also allows a direct comparison of the absolute localization performance across methods on each individual component. As shown in Fig.~\ref{fig:val1_allpro}, SPADE consistently attains the highest AUC-PRO@0.01 across all five components, with PaDiM forming a stable second tier on most of them. The performance gap is most pronounced on the sub-centimetre-scale components PZ4 and PZ5, where SPADE outperforms the other methods by the largest margin.

\subsection{Validation 2: Cross-Position Evaluation}
\label{subsec:val2}

Validation 2 addresses a practical data-scarcity condition: in industrial tactile inspection, acquiring nominal samples at every contact position can be costly, so few-shot target-position adaptation is relevant. The study trains on one acquisition position and evaluates cross-position generalization on PZ1 and PZ3, the components for which recurrent geometry makes position transfer meaningful. The training set is formed by fixing the fraction of \texttt{pos1} nominal samples at the stable value from Validation 1 and varying the fraction of \texttt{pos2} nominal samples from 0\% to 15\% in 5\% steps. The 0\% case is a zero-shot cross-position transfer; the remaining cases measure whether a small number of nominal target-position samples can recalibrate the normality model. Metrics are computed on an evaluation set containing nominal and defective samples from both \texttt{pos1} and \texttt{pos2}. Table~\ref{tab:val2_pro_paper} reports median AUROC, AUPRC, and AUC-PRO@0.01 over 10 seeds. A consistent pattern emerges across the three pixel-level metrics: moving from the zero-shot condition (\texttt{pos2}=0\%) to a small few-shot injection, already at 5\%, yields a clear improvement in most method--component pairs, while the gains from 5\% to 15\% are typically smaller. This suggests that a limited number of nominal samples from the second position is often sufficient to calibrate the detector to the acquisition shift. The main exceptions concern $\mathrm{AUC\text{-}PRO}@0.01$ on PZ3, where SPADE and InReaCh reach their best localization at 0\%. Under the strict $\mathrm{FPR}\le0.01$ regime, adding nominal variability can make the threshold more conservative and reduce defect-region coverage, even when AUROC or AUPRC improve. Across all injection levels, SPADE remains the strongest method in AUPRC and $\mathrm{AUC\text{-}PRO}@0.01$ on both PZ1 and PZ3, while PaDiM slightly exceeds it in AUROC on PZ3 at 10--15\%.

\begin{table*}[t]
\caption{Validation 2: median pixel-level metrics over 10 seeds at \texttt{pos2} $\in\{0\%,5\%,10\%,15\%\}$ nominal injection. \textbf{Bold}: row-wise maximum within each metric group; \underline{underline}: row-wise minimum.}
\label{tab:val2_pro_paper}
\centering
\scriptsize
\setlength{\tabcolsep}{3pt}
\renewcommand{\arraystretch}{0.95}
\resizebox{\textwidth}{!}{%
\begin{tabular}{ll rrrr | rrrr | rrrr}
\hline
& & \multicolumn{4}{c}{AUROC} & \multicolumn{4}{c}{AUPRC} & \multicolumn{4}{c}{AUC-PRO@0.01} \\
\cline{3-6}\cline{7-10}\cline{11-14}
\noalign{\vskip 1pt}
Method & Comp. & 0\% & 5\% & 10\% & 15\% & 0\% & 5\% & 10\% & 15\% & 0\% & 5\% & 10\% & 15\% \\
\hline
SPADE   & PZ1 & \underline{0.9562} & \textbf{0.9754} & 0.9753 & 0.9745 & \underline{0.3529} & 0.3749 & 0.3787 & \textbf{0.3832} & \underline{0.1286} & 0.1309 & 0.1306 & \textbf{0.1315} \\
PaDiM   & PZ1 & \underline{0.7752} & 0.9677 & 0.9705 & \textbf{0.9712} & \underline{0.0808} & 0.2978 & 0.3015 & \textbf{0.3025} & \underline{0.0314} & 0.1019 & 0.1017 & \textbf{0.1020} \\
FAPM    & PZ1 & \underline{0.7510} & 0.9620 & 0.9628 & \textbf{0.9632} & \underline{0.0340} & 0.2440 & 0.2446 & \textbf{0.2457} & \underline{0.0036} & \textbf{0.0648} & 0.0643 & 0.0639 \\
InReaCh & PZ1 & \underline{0.7869} & 0.9606 & 0.9618 & \textbf{0.9623} & \underline{0.0840} & 0.2378 & 0.2415 & \textbf{0.2423} & \underline{0.0273} & 0.0635 & 0.0633 & \textbf{0.0640} \\
\hline
SPADE   & PZ3 & \underline{0.9631} & \textbf{0.9820} & 0.9810 & 0.9809 & \underline{0.3712} & \textbf{0.3991} & 0.3967 & 0.3961 & \textbf{0.3397} & \underline{0.3310} & 0.3311 & 0.3316 \\
PaDiM   & PZ3 & \underline{0.9046} & 0.9804 & 0.9813 & \textbf{0.9818} & \underline{0.1030} & \textbf{0.3459} & 0.3448 & 0.3432 & \underline{0.0867} & \textbf{0.3054} & 0.2990 & 0.2968 \\
FAPM    & PZ3 & \underline{0.8880} & 0.9745 & 0.9748 & \textbf{0.9751} & \underline{0.0852} & \textbf{0.2556} & 0.2508 & 0.2497 & \underline{0.1019} & \textbf{0.2048} & 0.2037 & 0.2025 \\
InReaCh & PZ3 & \underline{0.9479} & 0.9748 & \textbf{0.9755} & 0.9754 & \underline{0.2748} & \textbf{0.2809} & 0.2793 & 0.2776 & \textbf{0.2258} & 0.2006 & 0.1990 & \underline{0.1983} \\
\hline
\end{tabular}%
}
\end{table*}

\subsection{Validation 3: Low- versus High-Resolution Comparison}
\label{subsec:val3}

Validation 3 compares $320\times240$ and $1280\times720$ data on PZ3, PZ4, and PZ5, the components for which both settings are available. This validation is particularly relevant from a design standpoint because, for very small components, it helps assess whether high-resolution tactile imaging is actually needed for defect localization or whether a lower-resolution, lower-cost sensing setup may be sufficient. The same within-position split as Validation 1 is used, and each method is trained with its stable fraction. High-resolution inference is performed with the tiled pipeline described in Section~\ref{sec:implementation}. As reported in Table~\ref{tab:val3_metrics_paper}, moving from LR to HR tiled acquisitions is generally beneficial for pixel-level localization, especially for AUPRC and $\mathrm{AUC\text{-}PRO}@0.01$, which emphasize precision under class imbalance and strict false-alarm control. This trend is particularly clear for PaDiM and FAPM, which improve consistently across PZ3--PZ5 and across all three pixel-level metrics. The method ranking also changes substantially: while SPADE is generally the strongest method in the LR setting, FAPM becomes the most competitive overall in HR, achieving the best HR AUROC on all three components and the best HR AUPRC/AUC-PRO@0.01 on most components. PaDiM also benefits markedly from HR and reaches the best HR AUC-PRO@0.01 on PZ3. These results suggest that the richer spatial information available in HR tactile data is particularly well exploited by methods based on distribution modeling and feature aggregation. In contrast, InReaCh shows the least stable LR-to-HR transfer, with a sharp degradation on PZ3 across all three metrics, making it the only method with a clearly inconsistent HR benefit in this validation.

\begin{table*}[t]
\caption{Validation 3: median pixel-level metrics over seeds 0--9 for low-resolution (LR) and high-resolution tiled (HR) inference. \textbf{Bold}: higher value within each LR/HR pair.}
\label{tab:val3_metrics_paper}
\centering
\small
\setlength{\tabcolsep}{5pt}
\renewcommand{\arraystretch}{1.0}
\begin{tabular}{ll rr | rr | rr}
\hline
& & \multicolumn{2}{c}{AUROC} & \multicolumn{2}{c}{AUPRC} & \multicolumn{2}{c}{AUC-PRO@0.01} \\
\cline{3-4}\cline{5-6}\cline{7-8}
\noalign{\vskip 1pt}
Method & Comp. & LR & HR & LR & HR & LR & HR \\
\hline
SPADE   & PZ3 & \textbf{0.9810} & 0.9603 & 0.3869 & \textbf{0.5125} & 0.3429 & \textbf{0.4706} \\
PaDiM   & PZ3 & 0.9783 & \textbf{0.9934} & 0.3269 & \textbf{0.4815} & 0.2986 & \textbf{0.5291} \\
FAPM    & PZ3 & 0.9709 & \textbf{0.9935} & 0.2519 & \textbf{0.5159} & 0.2084 & \textbf{0.5055} \\
InReaCh & PZ3 & \textbf{0.9704} & 0.8683 & \textbf{0.2784} & 0.0388 & \textbf{0.1929} & 0.0463 \\
\hline
SPADE   & PZ4 & \textbf{0.9925} & 0.9919 & 0.3229 & \textbf{0.4765} & 0.5717 & \textbf{0.7404} \\
PaDiM   & PZ4 & 0.9831 & \textbf{0.9965} & 0.1648 & \textbf{0.2372} & 0.2930 & \textbf{0.6388} \\
FAPM    & PZ4 & 0.9789 & \textbf{0.9979} & 0.1551 & \textbf{0.3950} & 0.2617 & \textbf{0.7676} \\
InReaCh & PZ4 & 0.9854 & \textbf{0.9896} & 0.2173 & \textbf{0.2390} & 0.3877 & \textbf{0.4880} \\
\hline
SPADE   & PZ5 & 0.9927 & \textbf{0.9953} & 0.1665 & \textbf{0.3033} & 0.3968 & \textbf{0.6143} \\
PaDiM   & PZ5 & 0.9907 & \textbf{0.9970} & 0.1266 & \textbf{0.2572} & 0.3202 & \textbf{0.5837} \\
FAPM    & PZ5 & 0.9857 & \textbf{0.9982} & 0.0874 & \textbf{0.3326} & 0.2462 & \textbf{0.6764} \\
InReaCh & PZ5 & 0.9871 & \textbf{0.9947} & 0.1051 & \textbf{0.2113} & 0.2676 & \textbf{0.4778} \\
\hline
\end{tabular}
\end{table*}

\section{Conclusion}
\label{sec:conclusion}

\begin{figure}[t]\vspace*{1pt}
\centering
\includegraphics[width=0.75\linewidth]{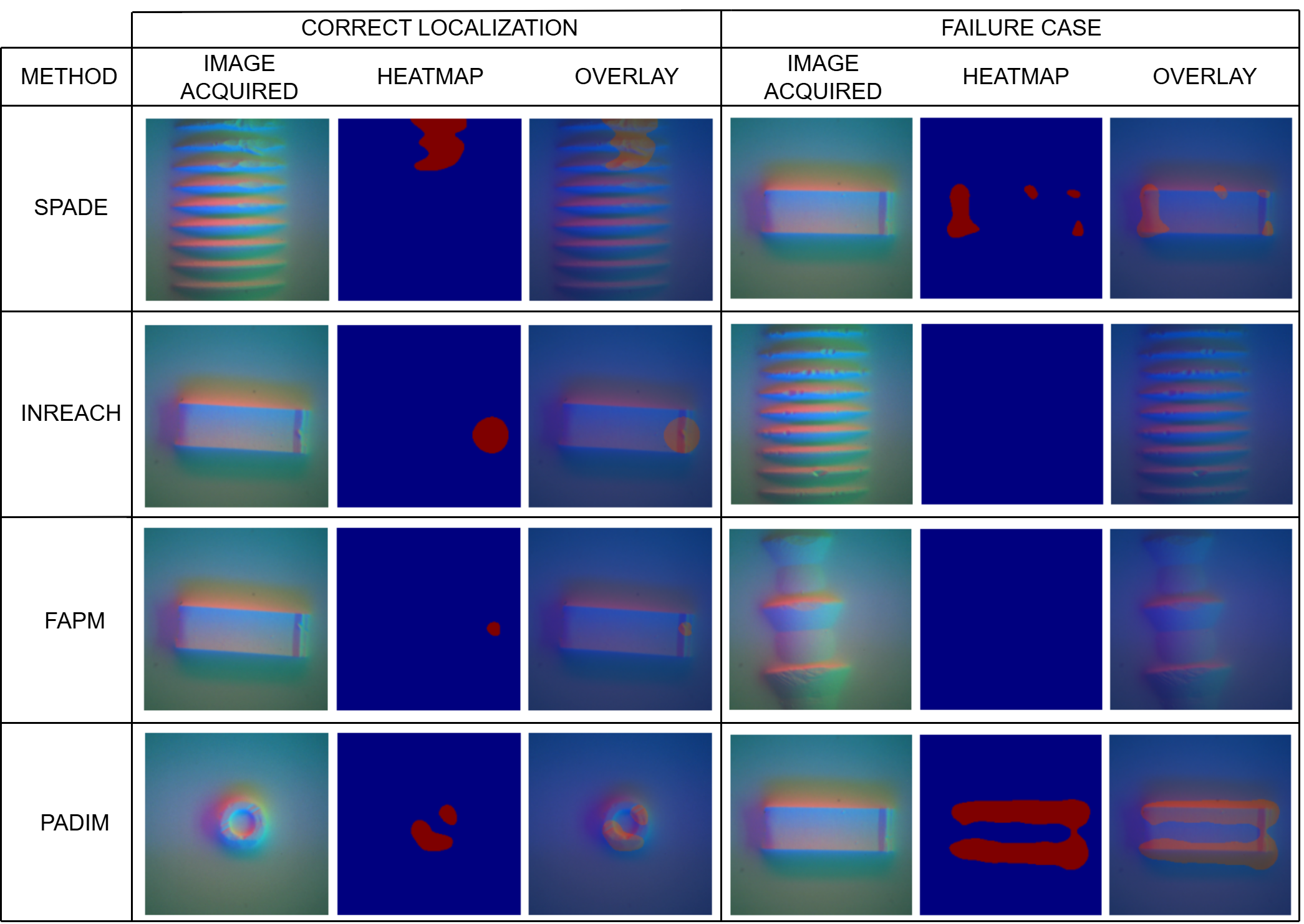}
\caption{Qualitative localization examples for the evaluated feature-embedding methods. Each image is a triptych reporting input, anomaly heatmap, and overlay.}
\label{fig:localization_examples}
\end{figure}

This work presents the first systematic evaluation of unsupervised feature-embedding methods for industrial anomaly detection on small components acquired with a vision-based tactile sensor. Table~\ref{tab:method_summary} summarizes the method behavior across the key deployment conditions; Fig.~\ref{fig:localization_examples} illustrates representative localization outcomes.
These results suggest that vision-based tactile sensing combined with unsupervised feature-embedding methods can represent a low-cost and flexible alternative to dedicated quality-control machines for small industrial components. No single method dominates across all conditions: SPADE offers the best localization and strongest zero-shot cross-position robustness but requires the full nominal pool, making it less suitable when gel wear must be minimized; InReaCh converges with the fewest contacts and handles position shifts without target samples, at the cost of lower absolute accuracy; PaDiM and FAPM benefit most from both few-shot cross-position injection and high-resolution acquisition, making them preferable when HR imaging is available and a small number of target-position samples can be collected.

\begin{table}[t]
\caption{Method comparison summary across deployment conditions. \checkmark\!\checkmark: strong advantage; \checkmark: advantage; $\circ$: moderate or mixed; $\times$: disadvantage.}
\label{tab:method_summary}
\vspace{3pt}
\centering
\scriptsize
\setlength{\tabcolsep}{2pt}
\renewcommand{\arraystretch}{0.85}
\resizebox{0.71\hsize}{!}{%
\begin{tabular}{lccccc}
\hline
 & \begin{tabular}[c]{@{}c@{}}Few nominal\\contacts\end{tabular}
 & \begin{tabular}[c]{@{}c@{}}LR\\localization\end{tabular}
 & \begin{tabular}[c]{@{}c@{}}Zero-shot\\cross-pos.\end{tabular}
 & \begin{tabular}[c]{@{}c@{}}Few-shot\\cross-pos.\ gain\end{tabular}
 & \begin{tabular}[c]{@{}c@{}}HR\\benefit\end{tabular} \\
\hline
SPADE   & $\times$               & \checkmark\!\checkmark & \checkmark\!\checkmark & $\circ$                & \checkmark \\
PaDiM   & \checkmark             & \checkmark             & $\circ$                & \checkmark\!\checkmark & \checkmark\!\checkmark \\
FAPM    & $\circ$                & $\circ$                & $\circ$                & \checkmark\!\checkmark & \checkmark\!\checkmark \\
InReaCh & \checkmark\!\checkmark & $\circ$                & \checkmark             & \checkmark             & $\times$ \\
\hline
\end{tabular}%
}
\end{table}

\section*{Acknowledgements}

The authors acknowledge \emph{Camozzi Research Center} for providing robotic equipment, as well as for their support.

\bibliographystyle{elsarticle-num}
\bibliography{paper_refs}

\begin{thebibliography}{10}
\expandafter\ifx\csname url\endcsname\relax
  \def\url#1{\texttt{#1}}\fi
\expandafter\ifx\csname urlprefix\endcsname\relax\def\urlprefix{URL }\fi
\expandafter\ifx\csname href\endcsname\relax
  \def\href#1#2{#2} \def\path#1{#1}\fi

\bibitem{cui2023_survey_unsupervised_industrial_images}
Y.~Cui, Z.~Liu, S.~Lian, A survey on unsupervised anomaly detection algorithms
  for industrial images, IEEE Access 11 (2023) 55297--55315.
\newblock \href {https://doi.org/10.1109/ACCESS.2023.3282993}
  {\path{doi:10.1109/ACCESS.2023.3282993}}.

\bibitem{tao2022_unsup_anom_localization_survey}
X.~Tao, X.~Gong, X.~Zhang, S.~Yan, C.~Adak, Deep learning for unsupervised
  anomaly localization in industrial images: A survey, IEEE Transactions on
  Instrumentation and Measurement 71 (2022) 1--21.
\newblock \href {https://doi.org/10.1109/TIM.2022.3196436}
  {\path{doi:10.1109/TIM.2022.3196436}}.

\bibitem{bergmann2021mvtec}
P.~Bergmann, K.~Batzner, M.~Fauser, D.~Sattlegger, C.~Steger, The {MVTec}
  anomaly detection dataset: A comprehensive real-world dataset for
  unsupervised anomaly detection, International Journal of Computer Vision 129
  (2021) 1038--1059.
\newblock \href {https://doi.org/10.1007/s11263-020-01400-4}
  {\path{doi:10.1007/s11263-020-01400-4}}.

\bibitem{lin2024_survey_rgb_3d_multimodal_uiad}
Y.~Lin, Y.~Chang, X.~Tong, J.~Yu, A.~Liotta, G.~Huang, W.~Song, D.~Zeng, Z.~Wu,
  Y.~Wang, W.~Zhang, A survey on rgb, 3d, and multimodal approaches for
  unsupervised industrial anomaly detection (2024).
\newblock \href {http://arxiv.org/abs/2410.21982} {\path{arXiv:2410.21982}}.

\bibitem{yuan2017gelsight}
W.~Yuan, S.~Dong, E.~H. Adelson, Gelsight: High-resolution robot tactile
  sensors for estimating geometry and force, Sensors 17~(12) (2017) 2762.
\newblock \href {https://doi.org/10.3390/s17122762}
  {\path{doi:10.3390/s17122762}}.

\bibitem{lambeta2020digit}
M.~Lambeta, P.-W. Chou, S.~Tian, B.~Yang, B.~Maloon, V.~R. Most, D.~Stroud,
  R.~Santos, A.~Byagowi, G.~Kammerer, D.~Jayaraman, R.~Calandra, {DIGIT}: A
  novel design for a low-cost compact high-resolution tactile sensor with
  application to in-hand manipulation, IEEE Robotics and Automation Letters
  5~(3) (2020) 3838--3845.
\newblock \href {https://doi.org/10.1109/LRA.2020.2977257}
  {\path{doi:10.1109/LRA.2020.2977257}}.

\bibitem{calandra2017feeling}
R.~Calandra, A.~Owens, M.~Upadhyaya, W.~Yuan, J.~Lin, E.~H. Adelson, S.~Levine,
  The feeling of success: Does touch sensing help predict grasp outcomes?, in:
  Proceedings of the 1st Annual Conference on Robot Learning, Vol.~78 of
  Proceedings of Machine Learning Research, PMLR, 2017, pp. 314--323.

\bibitem{han2025_vision_tactile_grasping}
Y.~Han, K.~Yu, R.~Batra, N.~Boyd, C.~Mehta, T.~Zhao, Y.~She, S.~Hutchinson,
  Y.~Zhao, Learning generalizable vision-tactile robotic grasping strategy for
  deformable objects via transformer, IEEE/ASME Transactions on Mechatronics
  30~(1) (2025) 554--566.
\newblock \href {https://doi.org/10.1109/TMECH.2024.3400789}
  {\path{doi:10.1109/TMECH.2024.3400789}}.

\bibitem{li2018_slip_detection}
J.~Li, S.~Dong, E.~H. Adelson, Slip detection with combined tactile and visual
  information, in: 2018 IEEE International Conference on Robotics and
  Automation (ICRA), 2018, pp. 7772--7777.
\newblock \href {https://doi.org/10.1109/ICRA.2018.8460495}
  {\path{doi:10.1109/ICRA.2018.8460495}}.

\bibitem{li2025vbtsreview}
H.~Li, Y.~Lin, C.~Lu, M.~Yang, E.~Psomopoulou, N.~F. Lepora, Classification of
  vision-based tactile sensors: A review, IEEE Sensors Journal (2025).
\newblock \href {http://arxiv.org/abs/2509.02478} {\path{arXiv:2509.02478}},
  \href {https://doi.org/10.1109/JSEN.2025.3599236}
  {\path{doi:10.1109/JSEN.2025.3599236}}.

\bibitem{mirzaee2025gelbelt}
M.~A. Mirzaee, H.-J. Huang, W.~Yuan, Gelbelt: A vision-based tactile sensor for
  continuous sensing of large surfaces, IEEE Robotics and Automation Letters
  (2025).
\newblock \href {http://arxiv.org/abs/2501.06263} {\path{arXiv:2501.06263}},
  \href {https://doi.org/10.1109/LRA.2025.3527306}
  {\path{doi:10.1109/LRA.2025.3527306}}.

\bibitem{zou2022spotdiff}
X.~Zou, Z.~Shi, Y.~Guo, J.~Ye, Spot-the-difference self-supervised pre-training
  for anomaly detection and segmentation, in: European Conference on Computer
  Vision (ECCV), 2022.

\bibitem{tabernik2019jim}
D.~Tabernik, S.~{\v{S}}ela, J.~Skvar{\v{c}}, D.~Sko{\v{c}}aj,
  Segmentation-based deep-learning approach for surface-defect detection,
  Journal of Intelligent ManufacturingAssociated dataset: Kolektor
  Surface-Defect Dataset (KolektorSDD) (2019).
\newblock \href {https://doi.org/10.1007/s10845-019-01476-x}
  {\path{doi:10.1007/s10845-019-01476-x}}.

\bibitem{yi2020_patch_svdd}
J.~Yi, S.~Yoon, Patch {SVDD}: Patch-level {SVDD} for anomaly detection and
  segmentation, in: Proceedings of the Asian Conference on Computer Vision
  (ACCV), 2020.

\bibitem{wang2021stfpm}
G.~Wang, S.~Han, E.~Ding, D.~Huang, Student-teacher feature pyramid matching
  for anomaly detection, arXiv preprint arXiv:2103.04257 (2021).
\newblock \href {http://arxiv.org/abs/2103.04257} {\path{arXiv:2103.04257}}.

\bibitem{defard2020padim}
T.~Defard, A.~Setkov, A.~Loesch, R.~Audigier, {PaDiM}: A patch distribution
  modeling framework for anomaly detection and localization (2020).
\newblock \href {http://arxiv.org/abs/2011.08785} {\path{arXiv:2011.08785}}.

\bibitem{cohen2020spade}
N.~Cohen, Y.~Hoshen, Sub-image anomaly detection with deep pyramid
  correspondences (2020).
\newblock \href {http://arxiv.org/abs/2005.02357} {\path{arXiv:2005.02357}}.

\bibitem{xu2022fapm}
J.~Xu, L.~Li, W.~Lv, Y.~Deng, Y.~Zhao, {FAPM}: Fast adaptive patch memory for
  real-time industrial anomaly detection (2022).
\newblock \href {http://arxiv.org/abs/2211.07381} {\path{arXiv:2211.07381}}.

\bibitem{mcintosh2023inreach}
L.~McIntosh, N.~Mahmood, Inter-realization channels: Unsupervised anomaly
  detection beyond one-class classification, in: Proceedings of the IEEE/CVF
  International Conference on Computer Vision (ICCV), 2023.

\bibitem{zavrtanik2021riad}
V.~Zavrtanik, M.~Kristan, D.~Sko{\v{c}}aj, Reconstruction by inpainting for
  visual anomaly detection, Pattern Recognition 112 (2021) 107706.
\newblock \href {https://doi.org/10.1016/j.patcog.2020.107706}
  {\path{doi:10.1016/j.patcog.2020.107706}}.

\bibitem{zhang2024dcae}
R.~Zhang, H.~Wang, M.~Feng, Y.~Liu, G.~Yang, Dual-constraint autoencoder and
  adaptive weighted similarity spatial attention for unsupervised anomaly
  detection, IEEE Transactions on Industrial Informatics 20~(7) (2024)
  9393--9403.
\newblock \href {https://doi.org/10.1109/TII.2024.3384583}
  {\path{doi:10.1109/TII.2024.3384583}}.

\bibitem{li2021cutpaste}
C.-L. Li, K.~Sohn, J.~Yoon, T.~Pfister, Cutpaste: Self-supervised learning for
  anomaly detection and localization (2021).
\newblock \href {http://arxiv.org/abs/2104.04015} {\path{arXiv:2104.04015}}.

\bibitem{zavrtanik2021draem}
V.~Zavrtanik, M.~Kristan, D.~Sko{\v{c}}aj, {DRAEM}: A discriminatively trained
  reconstruction embedding for surface anomaly detection (2021).
\newblock \href {http://arxiv.org/abs/2108.07610} {\path{arXiv:2108.07610}}.

\bibitem{cvat_zenodo}
{CVAT.ai Corporation},
  \href{https://zenodo.org/doi/10.5281/zenodo.3497105}{Computer vision
  annotation tool ({CVAT})}, Software repository and archival record, accessed:
  2026-02-08 (2026).
\newblock \href {https://doi.org/10.5281/zenodo.3497105}
  {\path{doi:10.5281/zenodo.3497105}}.
\newline\urlprefix\url{https://zenodo.org/doi/10.5281/zenodo.3497105}

\bibitem{rolih2024divideconquer}
B.~Rolih, D.~Ameln, A.~Vaidya, S.~Akcay, Divide and conquer: High-resolution
  industrial anomaly detection via memory efficient tiled ensemble (2024).
\newblock \href {http://arxiv.org/abs/2403.04932} {\path{arXiv:2403.04932}}.

\bibitem{bergmann2019mvtec}
P.~Bergmann, M.~Fauser, D.~Sattlegger, C.~Steger, The {MVTec} anomaly detection
  dataset: A comprehensive real-world dataset for unsupervised anomaly
  detection, in: Proceedings of the IEEE/CVF Conference on Computer Vision and
  Pattern Recognition (CVPR), 2019, pp. 9592--9600.
\newblock \href {https://doi.org/10.1109/CVPR.2019.00982}
  {\path{doi:10.1109/CVPR.2019.00982}}.

\end{thebibliography}

\end{document}